\documentclass[runningheads]{llncs}
\usepackage[T1]{fontenc}
\usepackage{graphicx,verbatim}
\usepackage{amsmath}

\usepackage{xurl}
\usepackage[hidelinks,breaklinks=true]{hyperref}
\newcommand\figref{Fig.~\ref}
\newcommand\tabref{Table~\ref}

\begin{document}

\title{PPIM: Pennes Physics-Informed Mamba for Heat-Source-Conditioned 3D Bioheat Simulation}
\titlerunning{PPIM for Heat-Source-Conditioned 3D Bioheat Simulation}

\author{Dongyun Lee\inst{1} \and
Kyungho Yoon\inst{2}\and Minwoo Shin\inst{1}\thanks{Corresponding author.}}
\authorrunning{D. Lee et al.}
\institute{Department of Software, Yonsei University (Mirae Campus), Wonju, 26493, Republic of Korea\and School of Mathematics and Computing, Yonsei University, Seoul, 03722, Republic of Korea\\
\email{\{leedy020923, mshin\}@yonsei.ac.kr}}

\maketitle

\begin{abstract}

    Three-dimensional bioheat simulation aims to predict transient temperature distributions in biological tissue and is commonly modeled using the Pennes bioheat equation, which combines thermal diffusion, perfusion-mediated heat loss, and external heat generation. In this study, we consider a controlled 3D Pennes bioheat simulation under a localized heat-source condition inspired by microwave ablation (MWA). To evaluate neural approximation performance, we compare three neural partial differential equation (PDE) solvers under the same controlled simulation: a spatial Fourier-feature physics-informed neural network (PINN), a generic PINNMamba temporal subsequence model, and Pennes Physics-Informed Mamba (PPIM). PPIM builds on the temporal subsequence model by incorporating conditioned heat-source input and Pennes-aware state-space model (SSM) decay initialization. All three neural models are trained under the same conditions with the same Pennes residual, and an explicit finite-difference method (FDM) solution is used only as the numerical reference. In a representative 600~s run, PPIM achieved the lowest MAE, relative $L_1$ error, and relative $L_2$ error among the evaluated neural solvers. Error maps further showed that the remaining PPIM errors were more concentrated near the heat-source region than across the rest of the domain. These results indicate that PPIM is effective for approximating the FDM reference final temperature field in this controlled simulation. The source code is available at \url{https://github.com/muvYun/PPIM}.
    
\end{abstract}

\section{Introduction}

    Three-dimensional bioheat simulation is important for predicting transient temperature distributions in biological tissue, particularly in thermal therapy settings. A representative example is microwave ablation (MWA), a minimally invasive thermal therapy widely used for the local treatment of liver tumors, where antenna-induced energy deposition produces localized tissue heating~\cite{brace2010microwave,hui2020microwave,simon2005microwave}. In this work, MWA is used as an application-inspired heating scenario for evaluating a computational component of thermal-therapy-oriented digital-twin modeling~\cite{shin2024phys,cho2026dt}, rather than as a full patient-specific treatment simulator.
    
    Tissue temperature dynamics are commonly modeled using the Pennes bioheat equation, which accounts for thermal diffusion, perfusion-mediated heat loss, and external heat generation~\cite{pennes1948analysis}. Pennes bioheat problems have traditionally been solved using numerical methods such as the finite-difference method (FDM)~\cite{leveque2007finite}. Here, we define a controlled 3D Pennes bioheat simulation problem with a localized heat-source condition inspired by MWA and evaluate neural partial differential equation (PDE) solvers against an explicit FDM reference.
    
    Recently, PINNs have been investigated as neural PDE solvers that learn from governing-equation residuals~\cite{raissi2019physics}, and PINN-based approaches have also been applied to Pennes bioheat problems~\cite{yesilyurt2026meshfree}. However, PINNs can exhibit failure modes under certain PDE conditions, and a reduction in residual loss does not always guarantee an accurate approximation of the solution~\cite{krishnapriyan2021failure}. Motivated by sequence-based physics-informed learning, PINNMamba has explored a state-space model (SSM)-based temporal-subsequence framework for PDE approximation~\cite{xu2025subsequential}.

    Building on this sequence-based formulation, we propose Pennes Physics-Informed Mamba (PPIM), a temporal-subsequence-based neural PDE solver for the controlled 3D Pennes bioheat simulation. PPIM incorporates two problem-specific components, conditioned heat-source input and Pennes-aware SSM decay initialization, to reflect the localized forcing field and representative diffusion--perfusion time scales of the Pennes equation.

    The main contributions of this paper are as follows:
    \begin{itemize}
        \item We define a controlled 3D Pennes bioheat simulation problem under an MWA-inspired localized heat-source condition.
        \item We propose PPIM, a temporal-subsequence-based neural PDE solver with conditioned heat-source input and Pennes-aware SSM decay initialization.
        \item We evaluate PINN, PINNMamba, and PPIM under the same FDM-referenced simulation condition using scalar error metrics and spatial error maps.
        \item PPIM achieves the lowest errors across all evaluated metrics.
    \end{itemize}

\section{Background}

    This study does not aim to reproduce a full clinical MWA simulation. Instead, it evaluates neural PDE solvers in a controlled 3D Pennes bioheat simulation under an MWA-inspired localized heat-source condition. For this controlled simulation, the heat source is modeled as a fixed specific absorption rate (SAR)-like Gaussian surrogate rather than a true electromagnetic SAR distribution. This simulation does not include full electromagnetic SAR modeling, temperature-dependent SAR redistribution, or large-vessel cooling~\cite{chiang2013computational,gorman2022numerical,ji2011expanded}.
    
    \subsection{Pennes Bioheat Simulation}

        We use the temperature elevation from the baseline body temperature,
        \(u(\mathbf{x},t)=T(\mathbf{x},t)-T_b\), as the predicted variable and solve the heat-capacity-normalized Pennes bioheat equation~\cite{pennes1948analysis}:
        \begin{equation}
        u_t-\alpha \nabla^2 u+\beta_{\mathrm{eff}}u-q=0.
        \end{equation}
        Here, \(\mathbf{x}=(x,y,z)\) denotes the spatial coordinate, \(t\) denotes time, \(T(\mathbf{x},t)\) is the tissue temperature, and \(T_b\) is the baseline body temperature. The term \(u_t\) denotes the temporal derivative of \(u\), and \(\nabla^2u=u_{xx}+u_{yy}+u_{zz}\) is the three-dimensional Laplacian.
    
        The coefficient \(C\) is the volumetric heat capacity, \(k\) is the thermal conductivity, and \(B\) is the nominal perfusion-related heat-loss coefficient. Accordingly, \(\alpha=k/C\) is the thermal diffusivity, \(\beta=B/C\) is the nominal heat-capacity-normalized perfusion coefficient, and \(\beta_{\mathrm{eff}}\) is the effective perfusion coefficient including temperature-dependent perfusion attenuation. \(Q_{\mathrm{ext}}\) denotes the external heat generation term, and \(q=Q_{\mathrm{ext}}/C\) is the heat-capacity-normalized external forcing. In this formulation, \(q\) represents the localized external heat source and does not include metabolic heat.
        
        The same forcing field \(q\) is used in the FDM computation and in the Pennes residual of all neural solvers. The resulting explicit FDM solution provides the numerical reference for evaluation~\cite{leveque2007finite}.
    
    \subsection{State-Space Sequence Modeling}
    
        An SSM represents temporal sequence information through the recurrent update of a latent state. In a Mamba-style SSM, the latent state for the \(\ell\)-th token in a temporal subsequence is updated using a state transition controlled by the decay matrix \(\mathbf{A}\):
        \begin{equation}
        \mathbf{h}_{\ell} = \exp(\Delta_{\ell}\mathbf{A})\mathbf{h}_{\ell-1} + \mathbf{b}_{\ell}.
        \end{equation}
        Here, \(\mathbf{h}_{\ell}\) denotes the latent state at the \(\ell\)-th token, and \(\mathbf{h}_{\ell-1}\) is the previous latent state. The index \(\ell\) denotes the token position within a temporal subsequence, \(\Delta_{\ell}\) is an input-dependent step size, and \(\mathbf{A}\) controls the state decay or memory time scale through the transition factor \(\exp(\Delta_{\ell}\mathbf{A})\). The term \(\mathbf{b}_{\ell}\) denotes the input-dependent additive update term that injects information from the current token into the latent state. It is distinct from the Pennes perfusion coefficient \(B\). Mamba introduces an input-dependent selective mechanism into this SSM structure, allowing temporal information to be selectively propagated or suppressed~\cite{gu2023mamba}.
    
        This SSM-based temporal sequence modeling is used by the sequence-based neural solvers considered in this study, including PINNMamba and PPIM~\cite{xu2025subsequential}. The model-specific architectures are described in the Method section.

\section{Method}

    \begin{figure}[t]
    \centering
    \includegraphics[width=\linewidth]{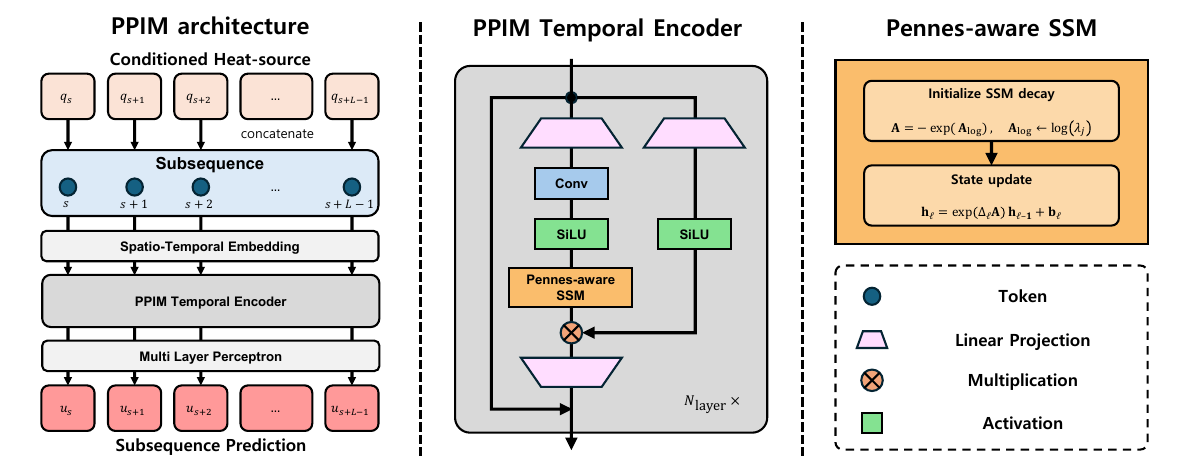}
    \caption{Overview of PPIM. Normalized spatio-temporal coordinates and \(q_{\mathrm{norm}}\) are concatenated to form temporal tokens, which are processed by the PPIM temporal encoder and mapped to temperature-elevation predictions.}
    \label{fig:PPIM_architecture}
    \end{figure}

    We propose PPIM, a neural PDE solver for three-dimensional Pennes bioheat simulations. PPIM follows the temporal subsequence and SSM-based sequence modeling of PINNMamba~\cite{xu2025subsequential}, while adding two problem-specific components: conditioned heat-source input and Pennes-aware SSM decay initialization. The overall architecture is shown in \figref{fig:PPIM_architecture}.

    \subsection{Conditioned Heat-Source Input}
    
        The localized forcing \(q(\mathbf{x},t)\) is prescribed by the simulation definition and corresponds to the heat-capacity-normalized external heat source in the Pennes residual. PPIM reuses this prescribed forcing field as an explicit input condition in addition to its role in residual evaluation. The normalized heat-source input is defined as
        \begin{equation}
        q_{\mathrm{norm}}(\mathbf{x},t) = \frac{q(\mathbf{x},t)}{q_{\mathrm{peak}}+\epsilon}.
        \end{equation}
        Here, \(q_{\mathrm{norm}}(\mathbf{x},t)\) is the dimensionless normalized heat-source feature, \(q_{\mathrm{peak}}\) is the peak value of the Gaussian heat source, and \(\epsilon\) is a small positive constant (e.g., \(\epsilon=10^{-12}\)) used for numerical stability. At each subsequence time step, \(q_{\mathrm{norm}}(\mathbf{x},t)\) is concatenated with the normalized spatio-temporal coordinate feature \((\bar{x},\bar{y},\bar{z},\bar{t})\) to form the input token. This allows PPIM to condition its prediction on the location and relative intensity of the localized heat source. Since \(q_{\mathrm{norm}}\) is derived from the prescribed source rather than from the FDM solution, it does not introduce additional target labels or supervised data.
    
    \subsection{Pennes-aware SSM Decay Initialization}
    
        In an SSM, the state decay controls the memory time scale of the temporal representation through the transition term \(\exp(\Delta_{\ell}\mathbf{A})\mathbf{h}_{\ell-1}\). Here, \(\Delta_{\ell}\) is the input-dependent step size for the \(\ell\)-th token, \(\mathbf{A}\) is the SSM decay matrix, and \(\mathbf{h}_{\ell-1}\) is the previous latent state. Instead of using a generic decay initialization, PPIM initializes the SSM state decay using representative diffusion--perfusion time scales implied by the Pennes bioheat equation.
        
        Let \(\mu_j\) denote a representative Laplacian spectral scale for the \(j\)-th decay component. The index \(j\) enumerates the initialized SSM decay scales. To reflect the range of perfusion attenuation, PPIM uses two decay families:
        \begin{equation}
        \lambda_j^{\mathrm{off}} = \alpha\mu_j, \qquad \lambda_j^{\mathrm{on}} = \alpha\mu_j+\beta, \qquad \mathbf{A}_{\log} \leftarrow \log(\lambda_j).
        \end{equation}
        Here, \(\alpha\) is the thermal diffusivity and \(\beta\) is the nominal perfusion coefficient defined in the Pennes formulation. The term \(\lambda_j^{\mathrm{off}}\) represents a diffusion-dominant decay rate, while \(\lambda_j^{\mathrm{on}}\) represents diffusion with perfusion-mediated decay. The symbol \(\lambda_j\) denotes an initialized decay rate selected from these two decay families.
    
        Through the Mamba parameterization \(\mathbf{A}=-\exp(\mathbf{A}_{\log})\) ~\cite{gu2023mamba}, this initialization determines the initial decay matrix \(\mathbf{A}\) used in the SSM state transition. The parameter \(\mathbf{A}_{\log}\) denotes the trainable logarithmic decay parameter of the SSM. The initialized decay parameters remain trainable during optimization, and the proposed initialization is used only as a physics-informed prior rather than as an exact spectral solver for the full temperature-dependent Pennes problem.

\section{Experiments}

    \subsection{Simulation Setup}

        We compare three neural solvers under the same controlled 3D Pennes bioheat simulation condition: a spatial Fourier-feature PINN~\cite{tancik2020fourier}, a generic PINNMamba temporal subsequence model, and PPIM. All neural models use the same Pennes equation, MWA-inspired Gaussian heat-source distribution, material properties, initial condition, boundary condition, and Pennes residual. The FDM solution is used only as the numerical reference and not as supervised training data~\cite{leveque2007finite}.
    
        Evaluation is performed on the final temperature field at \(t=600~\mathrm{s}\) using mean absolute error (MAE), relative \(L_1\) error, and relative \(L_2\) error against the FDM reference. In addition to scalar metrics, we visualize the final heat distribution, signed error, absolute error, relative \(L_1\) contribution, and relative \(L_2\) contribution maps to analyze the spatial distribution of approximation errors.
    
    \subsection{Heat-Source and Material Settings}
     
        The heat-source condition is defined as a localized SAR-like Gaussian surrogate around a \(15~\mathrm{mm}\) active segment, instead of an antenna-specific SAR field computed from a full electromagnetic simulation. This choice is intentional: prior computational MWA studies indicate that SAR-based modeling typically requires electromagnetic field simulation and may involve temperature-dependent dielectric or tissue-property modeling, which would introduce additional modeling factors beyond the neural PDE solver comparison considered here~\cite{chiang2013computational,gorman2022numerical,ji2011expanded}. Using a fixed Gaussian surrogate keeps the external forcing field \(q\) prescribed and identical for the FDM reference and neural PDE solvers.
        
        The source protocol is inspired by a live porcine liver MWA study that reported a \(65~\mathrm{W}\), \(600~\mathrm{s}\) protocol with a \(15~\mathrm{mm}\) active radiating segment~\cite{hui2020microwave}. The generator power is mapped to the Gaussian surrogate using a fixed effective source-power factor of \(0.64\), giving \(41.6~\mathrm{W}\). This factor is used as a surrogate source-power calibration rather than as a direct measurement of tissue absorption. It was chosen to be close to the reported delivered-power estimate and to remain consistent with the applied/effective tissue-power interpretation used in MWA power modeling~\cite{dong2017applied,hui2020microwave}.
    
        The liver thermal parameters are fixed as \(C=3.685~\mathrm{J/(cm^3\,^\circ C)}\), \(k=0.0055~\mathrm{W/(cm\,^\circ C)}\), and \(B=0.05~\mathrm{W/(cm^3\,^\circ C)}\), following Pennes-based hepatic thermal modeling settings~\cite{pennes1948analysis,prakash2010theoretical}. These values define \(\alpha=k/C\) and \(\beta=B/C\) in the heat-capacity-normalized Pennes equation. The same source and material settings are used for the FDM reference and all neural solvers.
    
    \subsection{Training Details}
    
        All neural models are trained using the same physics-informed training loop. A hard constraint transform is used to satisfy \(u=0\) at \(t=0\) and \(u=0\) on the domain boundary~\cite{lagaris1998artificial}. Therefore, no additional soft initial or boundary condition losses are used. The physics loss is computed from the same discrete Pennes residual, and all neural models are trained for 3000 iterations using Adam~\cite{kingma2014adam} with the same learning rate. Source-aware sampling is applied to avoid undersampling the localized Gaussian heat-source region.

        The PINN is a non-sequential model and is therefore trained without sequence consistency loss. In contrast, PINNMamba and PPIM use temporal subsequences of length \(L=7\) and include the subsequence consistency loss. PPIM uses the same temporal subsequence training loop as PINNMamba, with the \(q_{\mathrm{norm}}\) input and Pennes-aware SSM decay initialization described in the Method section.
    
   \subsection{Results}

        \begin{table}[t!]
        \caption{Comparison of neural solvers against the FDM reference at \(t=600~\mathrm{s}\).}
        \label{tab:neural_solver_results}
        \centering
        \begin{tabular*}{0.82\linewidth}{@{\extracolsep{\fill}}cccc}
        \hline
        Model & MAE \((^\circ\mathrm{C})\) & Rel. \(L_1\) Error & Rel. \(L_2\) Error \\
        \hline
        PINN & 0.1999 & 0.2452 & 0.1135 \\
        PINNMamba & 0.1067 & 0.1309 & 0.0719 \\
        \textbf{PPIM} & \textbf{0.0692} & \textbf{0.0849} & \textbf{0.0465} \\
        \hline
        \end{tabular*}
        \end{table}

        \begin{figure}[t!]
        \centering
        \includegraphics[width=0.93\textwidth]{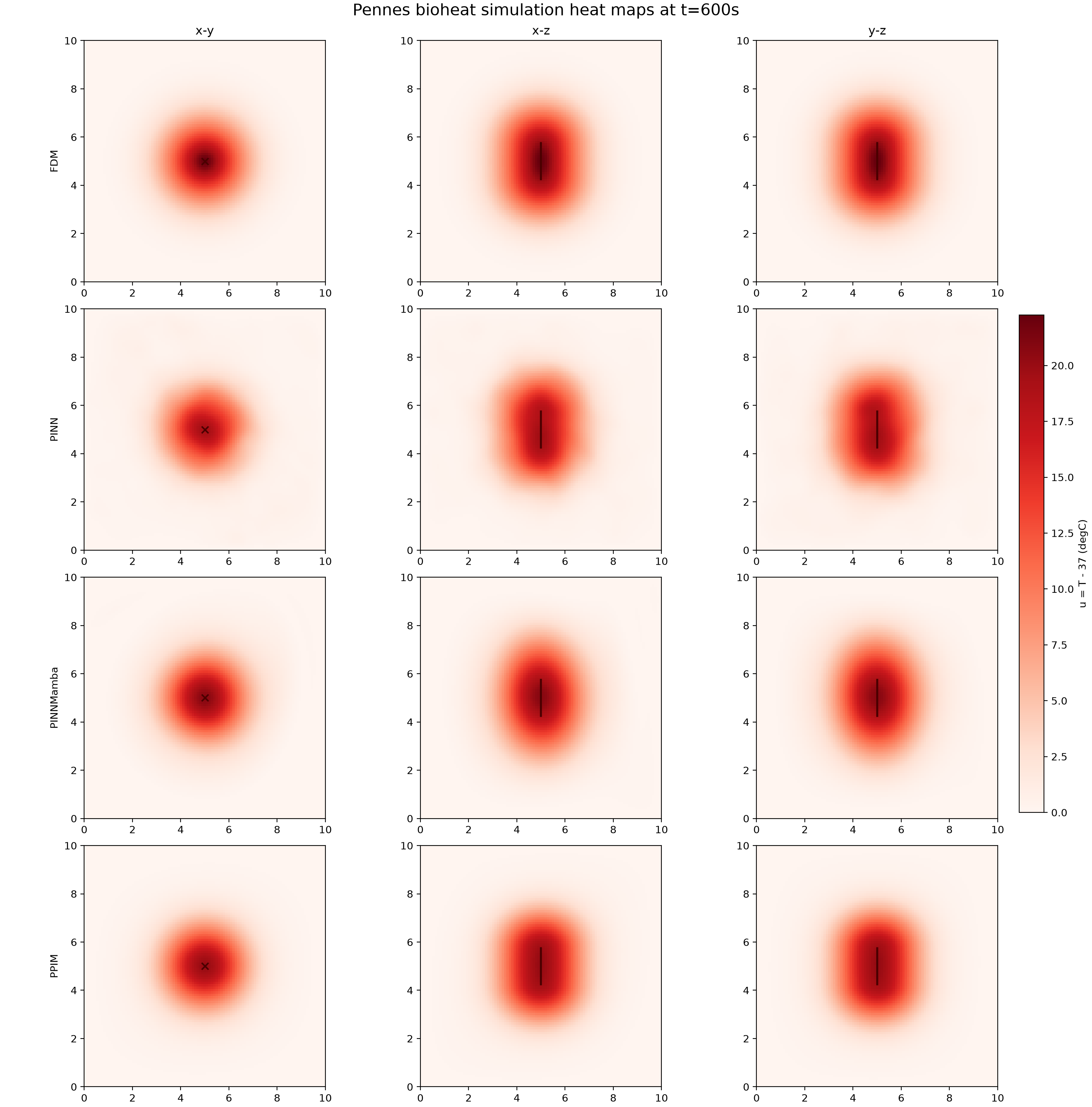}
        \caption{Final heat distributions at \(t=600~\mathrm{s}\). The field represents temperature elevation \(u=T-37~^\circ\mathrm{C}\).}
        \label{fig:heat_distribution_maps}
        \end{figure}
    
        \tabref{tab:neural_solver_results} shows that PINN has the highest errors, PINNMamba produces intermediate errors, and PPIM achieves the lowest errors across MAE, relative \(L_1\), and relative \(L_2\). \figref{fig:heat_distribution_maps} shows that all neural solvers form a localized heating pattern around the antenna-like source, while PPIM more closely follows the FDM reference temperature field and surrounding decay.

        \begin{figure}[t!]
        \centering
        \begin{minipage}{0.49\linewidth}
            \centering
            \includegraphics[width=\linewidth]{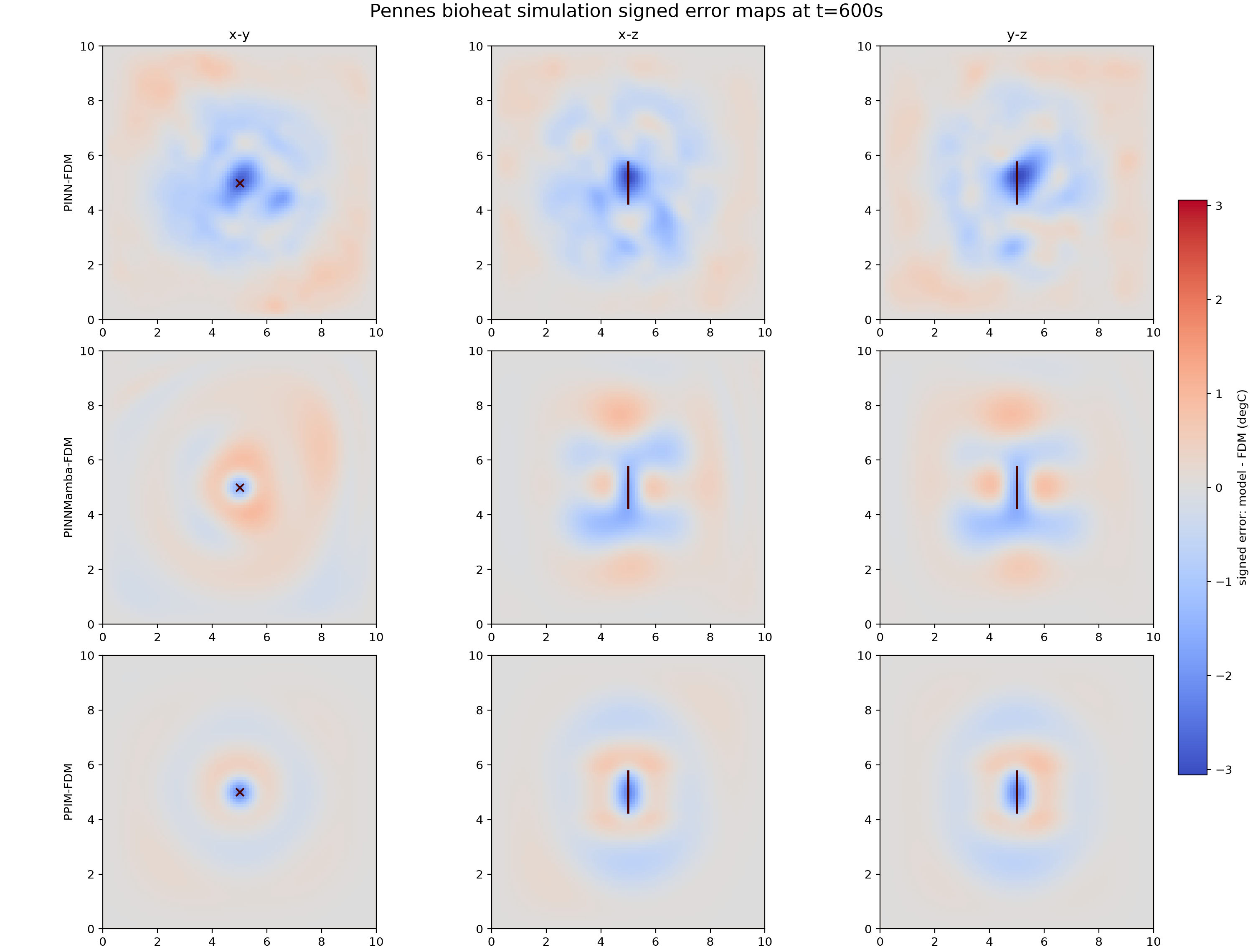}
            \vspace{1mm}
            \footnotesize (a) Signed error maps
        \end{minipage}
        \hfill
        \begin{minipage}{0.49\linewidth}
            \centering
            \includegraphics[width=\linewidth]{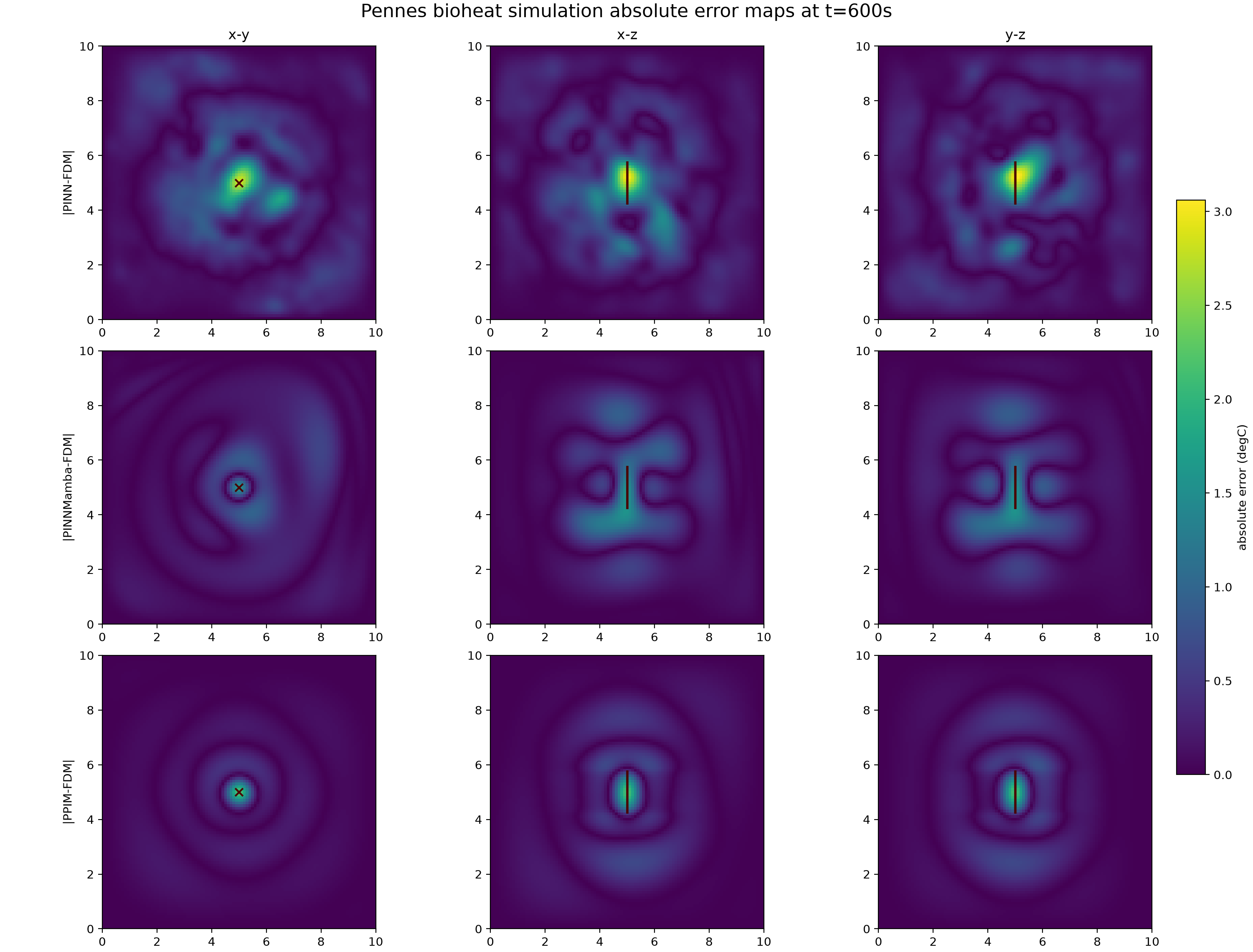}
            \vspace{1mm}
            \footnotesize (b) Absolute error maps
        \end{minipage}
        \caption{Signed and absolute error maps at \(t=600~\mathrm{s}\). Panel (a) shows \(u_\theta-u_{\mathrm{FDM}}\), and panel (b) shows \(|u_\theta-u_{\mathrm{FDM}}|\).}
        \label{fig:error_maps}
        \end{figure}
    
        The signed and absolute error maps in \figref{fig:error_maps} show that PINN errors are broadly distributed around the source and transition regions. PINNMamba reduces this spatial spread, and PPIM produces the most spatially confined error pattern among the neural solvers.

        \begin{figure}[t!]
        \centering
        \begin{minipage}{0.49\linewidth}
            \centering
            \includegraphics[width=\linewidth]{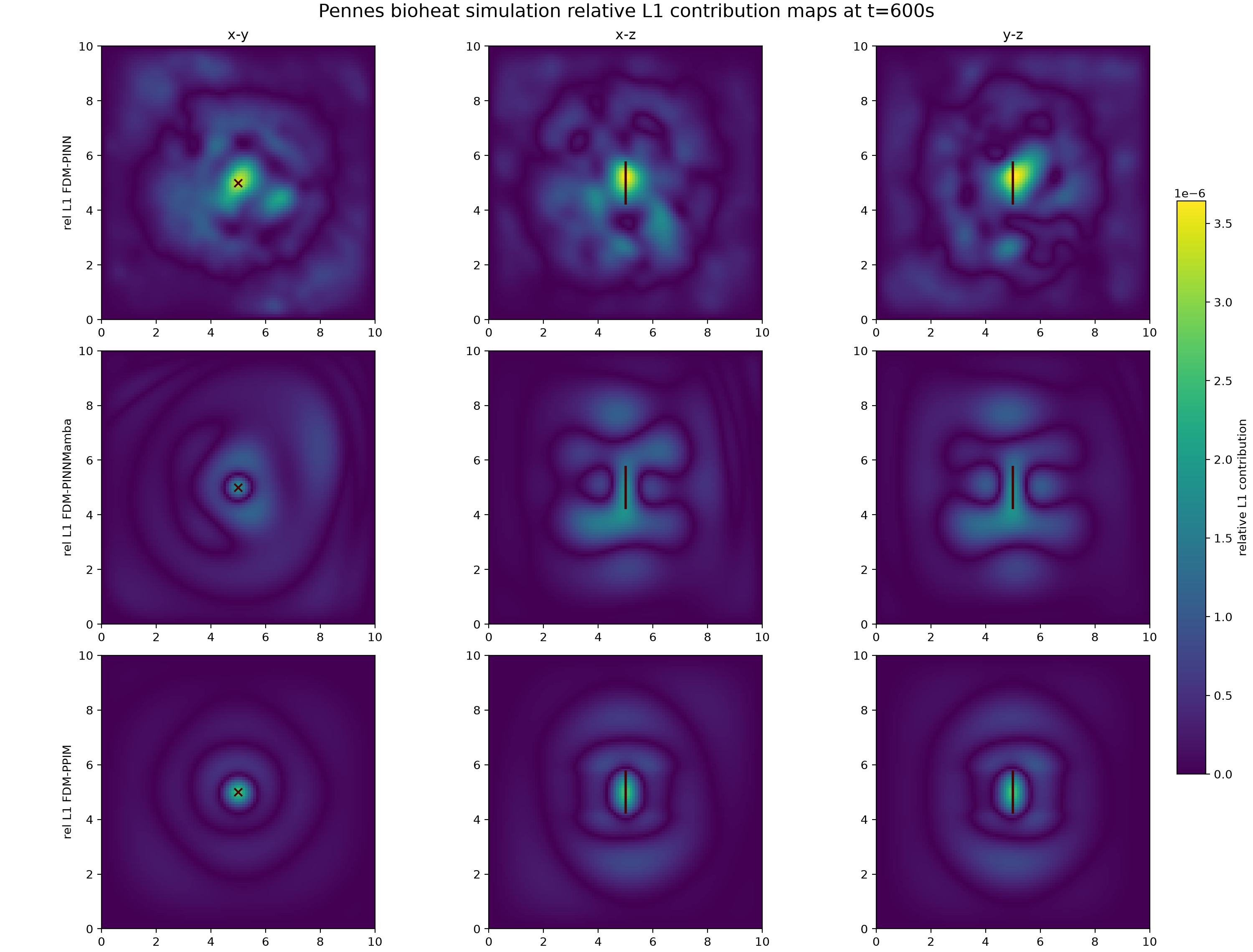}
            \vspace{1mm}
            \footnotesize (a) Relative \(L_1\) contribution
        \end{minipage}
        \hfill
        \begin{minipage}{0.49\linewidth}
            \centering
            \includegraphics[width=\linewidth]{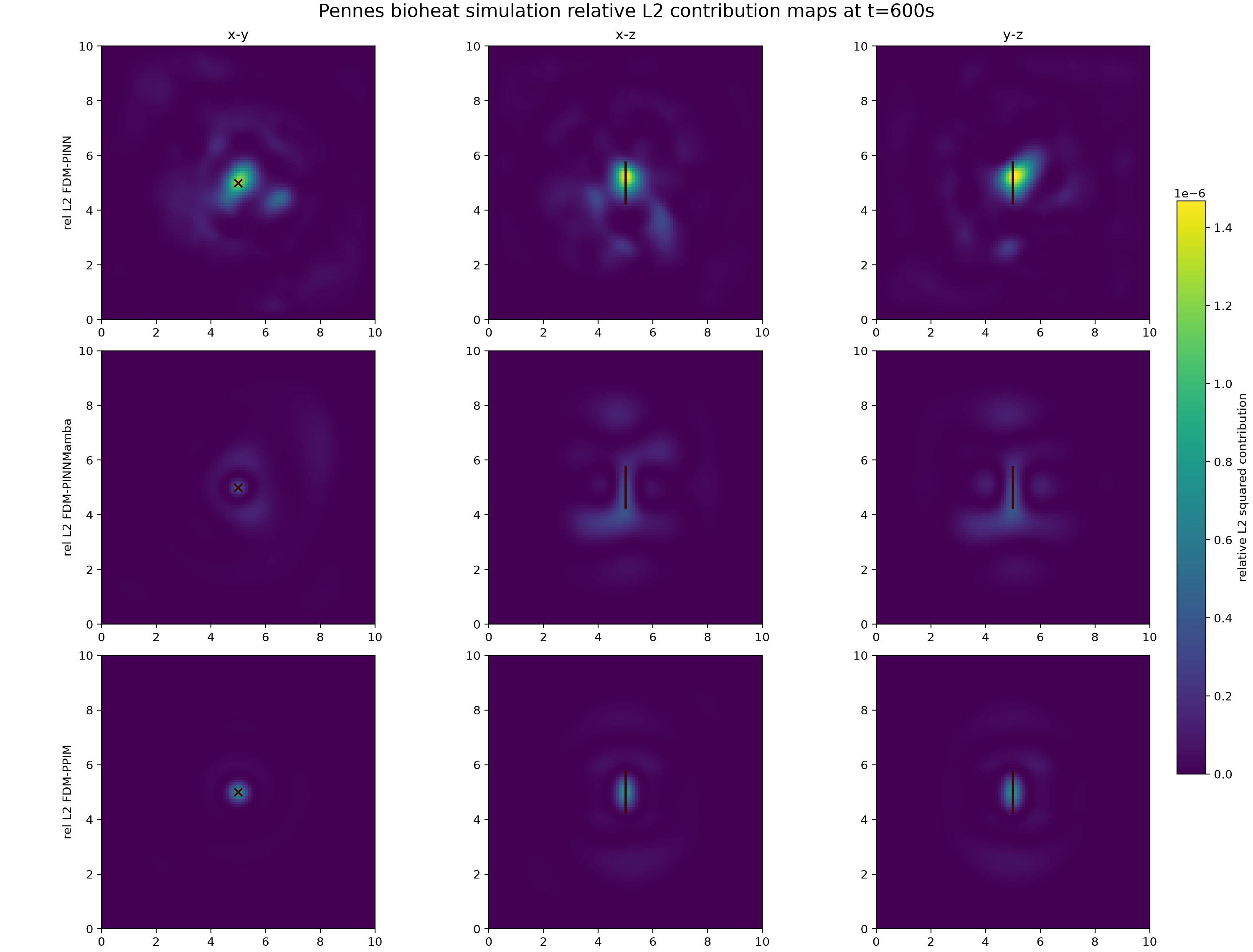}
            \vspace{1mm}
            \footnotesize (b) Relative \(L_2\) contribution
        \end{minipage}
        \caption{Relative error contribution maps at \(t=600~\mathrm{s}\). Panel (a) shows the pointwise normalized absolute-error contribution, and panel (b) shows the pointwise normalized squared-error contribution.}
        \label{fig:relative_error_maps}
        \end{figure}
    
        \figref{fig:relative_error_maps} shows the spatial contributions to the relative \(L_1\) and \(L_2\) metrics. The contribution maps follow the same ordering as the scalar errors: PINN shows the largest source-adjacent contribution, PINNMamba reduces it, and PPIM gives the lowest and most localized contribution.

        Together, the scalar metrics and spatial analyses indicate that PPIM provides the closest neural approximation to the FDM reference among the evaluated neural PDE solvers.

\section{Conclusion}

    In this study, we proposed PPIM, a temporal-subsequence-based neural PDE solver for a controlled 3D Pennes bioheat simulation under an MWA-inspired localized heat-source condition. PPIM follows the subsequence and SSM-based sequence modeling of PINNMamba~\cite{xu2025subsequential}, while incorporating conditioned heat-source input and Pennes-aware SSM decay initialization.

    We evaluated PINN, PINNMamba, and PPIM under the same controlled 3D Pennes bioheat simulation condition, using the FDM solution as the numerical reference. In the representative 600~s run, PPIM achieved the lowest MAE, relative \(L_1\), and relative \(L_2\) errors among the evaluated neural solvers. The heat distribution and error maps also showed that PPIM had the closest spatial agreement with the reference.
    
    These results indicate that PPIM is effective for approximating the reference final temperature field in a controlled 3D Pennes bioheat simulation.

\begin{credits}
\subsubsection{\ackname} 
This work was supported by the Korea Medical Device Development Foundation grant funded by the Korean government (the Ministry of Science and ICT, the Ministry of Trade, Industry and Energy, the Ministry of Health and Welfare, and the Ministry of Food and Drug Safety) (Grant No. RS-2026-25543484).

This research was also supported by the ANCHOR Program through the Gangwon ANCHOR Center, funded by the Ministry of Education (MOE) and Gangwon State (G.S.), Republic of Korea (Grant No. 2026-ANCHOR-10-006).

This research was further supported by the Ministry of Science and ICT (MSIT), Korea, under the National Program in Medical AI Semiconductor (Grant No. 2024-0-00096), supervised by the Institute of Information \& Communications Technology Planning \& Evaluation (IITP) in 2026.

\subsubsection{\discintname}
The authors have no competing interests to declare that are relevant to the content of this article.
\end{credits}

\bibliographystyle{splncs04}
\bibliography{Paper-12}

@article{simon2005microwave,
  title = {{Microwave Ablation: Principles and Applications}},
  author = {Simon, Caroline J. and Dupuy, Damian E. and Mayo-Smith, William W.},
  journal = {RadioGraphics},
  volume = {25},
  number = {suppl 1},
  pages = {S69--S83},
  year = {2005},
  doi = {10.1148/rg.25si055501}
}

@article{brace2010microwave,
  title = {{Microwave Tissue Ablation: Biophysics, Technology, and Applications}},
  author = {Brace, Christopher L.},
  journal = {Critical Reviews in Biomedical Engineering},
  volume = {38},
  number = {1},
  pages = {65--78},
  year = {2010},
  doi = {10.1615/CritRevBiomedEng.v38.i1.60}
}

@article{hui2020microwave,
  title = {{Microwave ablation of the liver in a live porcine model: the impact of power, time and total energy on ablation zone size and shape}},
  author = {Hui, Terrence Chi Hong and Brace, Christopher Lee and Hinshaw, J. Louis and Quek, Lawrence Han Hwee and Huang, Ivan Kuang Hsin and Kwan, Justin and Lim, Gavin Hock Tai and Lee, Jr, Fred T and Pua, Uei},
  journal = {International Journal of Hyperthermia},
  volume = {37},
  number = {1},
  pages = {668--676},
  year = {2020},
  publisher = {Taylor \& Francis},
  doi = {10.1080/02656736.2020.1774083}
}

@article{pennes1948analysis,
  title = {{Analysis of Tissue and Arterial Blood Temperatures in the Resting Human Forearm}},
  author = {Pennes, Harry H.},
  journal = {Journal of Applied Physiology},
  volume = {1},
  number = {2},
  pages = {93--122},
  year = {1948},
  doi = {10.1152/jappl.1948.1.2.93}
}

@article{prakash2010theoretical,
  title = {{Theoretical Modeling for Hepatic Microwave Ablation}},
  author = {Prakash, Punit},
  journal = {The Open Biomedical Engineering Journal},
  volume = {4},
  pages = {27--38},
  year = {2010},
  doi = {10.2174/1874120701004010027}
}

@article{chiang2013computational,
  title = {{Computational Modelling of Microwave Tumour Ablations}},
  author = {Chiang, Jason and Wang, Peng and Brace, Christopher L.},
  journal = {International Journal of Hyperthermia},
  volume = {29},
  number = {4},
  pages = {308--317},
  year = {2013},
  publisher = {Taylor \& Francis},
  doi = {10.3109/02656736.2013.799295}
}

@article{ji2011expanded,
  title = {{Expanded Modeling of Temperature-Dependent Dielectric Properties for Microwave Thermal Ablation}},
  author = {Ji, Zhen and Brace, Christopher L.},
  journal = {Physics in Medicine and Biology},
  volume = {56},
  number = {16},
  pages = {5249--5264},
  year = {2011},
  publisher = {IOP Publishing},
  doi = {10.1088/0031-9155/56/16/011}
}

@article{gorman2022numerical,
  title = {{Numerical Simulation of Microwave Ablation in the Human Liver}},
  author = {Gorman, John and Tan, Winston and Abraham, John},
  journal = {Processes},
  volume = {10},
  number = {2},
  pages = {361},
  year = {2022},
  publisher = {MDPI},
  doi = {10.3390/pr10020361}
}

@book{leveque2007finite,
  title = {{Finite Difference Methods for Ordinary and Partial Differential Equations: Steady-State and Time-Dependent Problems}},
  author = {LeVeque, Randall J.},
  publisher = {Society for Industrial and Applied Mathematics},
  address = {Philadelphia, PA},
  year = {2007},
  isbn = {978-0-898716-29-0},
  doi = {10.1137/1.9780898717839}
}

@article{raissi2019physics,
  title = {{Physics-Informed Neural Networks: A Deep Learning Framework for Solving Forward and Inverse Problems Involving Nonlinear Partial Differential Equations}},
  author = {Raissi, Maziar and Perdikaris, Paris and Karniadakis, George Em},
  journal = {Journal of Computational Physics},
  volume = {378},
  pages = {686--707},
  year = {2019},
  publisher = {Elsevier},
  doi = {10.1016/j.jcp.2018.10.045}
}

@article{yesilyurt2026meshfree,
  title = {{Mesh-Free Modeling of Heat Transfer Dynamics for Rapid Assessment of Necrosis Zones in Hepatic Tumor Radiofrequency Ablation Using Physics-Informed Neural Networks}},
  author = {Ye{\c{s}}ilyurt, Muhammet Kaan},
  journal = {Acadlore Transactions on AI and Machine Learning},
  volume = {5},
  number = {1},
  pages = {73--88},
  year = {2026},
  doi = {10.56578/ataiml050107}
}

@inproceedings{krishnapriyan2021failure,
  title = {{Characterizing Possible Failure Modes in Physics-Informed Neural Networks}},
  author = {Krishnapriyan, Aditi S. and Gholami, Amir and Zhe, Shandian and Kirby, Robert M. and Mahoney, Michael W.},
  booktitle = {Advances in Neural Information Processing Systems},
  volume = {34},
  pages = {26548--26560},
  year = {2021},
  url = {https://proceedings.neurips.cc/paper/2021/hash/df438e5206f31600e6ae4af72f2725f1-Abstract.html}
}

@inproceedings{xu2025subsequential,
  title = {{Sub-Sequential Physics-Informed Learning with State Space Model}},
  author = {Xu, Chenhui and Liu, Dancheng and Hu, Yuting and Li, Jiajie and Qin, Ruiyang and Zheng, Qingxiao and Xiong, Jinjun},
  booktitle = {Proceedings of the 42nd International Conference on Machine Learning},
  editor = {Singh, Aarti and Fazel, Maryam and Hsu, Daniel and Lacoste-Julien, Simon and Berkenkamp, Felix and Maharaj, Tegan and Wagstaff, Kiri and Zhu, Jerry},
  series = {Proceedings of Machine Learning Research},
  volume = {267},
  pages = {69507--69525},
  publisher = {PMLR},
  month = jul,
  year = {2025},
  note = {13--19 Jul 2025},
  url = {https://proceedings.mlr.press/v267/xu25t.html}
}

@misc{gu2023mamba,
  title = {{Mamba: Linear-Time Sequence Modeling with Selective State Spaces}},
  author = {Gu, Albert and Dao, Tri},
  year = {2023},
  eprint = {2312.00752},
  archivePrefix = {arXiv},
  primaryClass = {cs.LG},
  doi = {10.48550/arXiv.2312.00752}
}

@inproceedings{tancik2020fourier,
  title = {{Fourier Features Let Networks Learn High Frequency Functions in Low Dimensional Domains}},
  author = {Tancik, Matthew and Srinivasan, Pratul P. and Mildenhall, Ben and Fridovich-Keil, Sara and Raghavan, Nithin and Singhal, Utkarsh and Ramamoorthi, Ravi and Barron, Jonathan T. and Ng, Ren},
  booktitle = {Advances in Neural Information Processing Systems},
  volume = {33},
  pages = {7537--7547},
  publisher = {Curran Associates, Inc.},
  year = {2020},
  url = {https://proceedings.neurips.cc/paper/2020/hash/55053683268957697aa39fba6f231c68-Abstract.html}
}

@inproceedings{dong2017applied,
  title = {{Study of Applied Tissue Power in Microwave Ablation}},
  author = {Dong, Tong and Nan, Qun and Tian, Zhen and Nie, Xiaohui and Cheng, Yanyan},
  booktitle = {Proceedings of the International Conference on Computational Methods},
  year = {2017},
  note = {{ICCM 2017}},
  url = {https://www.sci-en-tech.com/ICCM2017/PDFs/2250-9628-1-PB.pdf}
}

@article{lagaris1998artificial,
  title = {{Artificial Neural Networks for Solving Ordinary and Partial Differential Equations}},
  author = {Lagaris, Isaac E. and Likas, Aristidis and Fotiadis, Dimitrios I.},
  journal = {IEEE Transactions on Neural Networks},
  volume = {9},
  number = {5},
  pages = {987--1000},
  year = {1998},
  publisher = {IEEE},
  doi = {10.1109/72.712178}
}

@misc{kingma2014adam,
  title = {{Adam: A Method for Stochastic Optimization}},
  author = {Kingma, Diederik P. and Ba, Jimmy},
  year = {2014},
  eprint = {1412.6980},
  archivePrefix = {arXiv},
  primaryClass = {cs.LG},
  doi = {10.48550/arXiv.1412.6980}
}

@article{shin2024phys,
  title = {{{PhysRFANet}: Physics-Guided Neural Network for Real-Time Prediction of Thermal Effect during Radiofrequency Ablation Treatment}},
  author = {Shin, Minwoo and Seo, Minjee and Cho, Seonaeng and Park, Juil and Kwon, Joon Ho and Lee, Deukhee and Yoon, Kyungho},
  journal = {Engineering Applications of Artificial Intelligence},
  volume = {138},
  pages = {109349},
  year = {2024},
  issn = {0952-1976},
  doi = {10.1016/j.engappai.2024.109349}
}

@inproceedings{cho2026dt,
  title = {{Towards Digital Twin of {RF} Ablation: Real-Time Prediction of Time-Dependent Thermal Effects Using Transformer}},
  author = {Cho, Seonaeng and Seo, Minjee and Shin, Minwoo and Yoon, Kyungho},
  editor = {Li, Lei and Jirsa, Viktor and Feng, Jianfeng and Deng, Jun and Dede', Luca and An, Sora and Lyu, Yilin and Liu, Xiaoyue},
  booktitle = {Digital Twin for Healthcare},
  series = {Lecture Notes in Computer Science},
  volume = {16193},
  pages = {69--78},
  publisher = {Springer Nature Switzerland},
  address = {Cham},
  year = {2026},
  isbn = {978-3-032-07694-6},
  doi = {10.1007/978-3-032-07694-6_7}
}

\end{document}